\documentclass[letterpaper, 10 pt, conference]{ieeeconf}  

\IEEEoverridecommandlockouts                              

\usepackage[table]{xcolor}
\definecolor{awesome}{rgb}{1.0, 0.13, 0.32}

\usepackage{graphics} 
\usepackage{epsfig} 
\usepackage[color=red]{todonotes}
\usepackage{mathptmx} 
\usepackage{times} 
\usepackage{amsmath} 
\usepackage{amssymb}  
\usepackage{microtype}
\usepackage{multirow}
\usepackage{vcell}
\usepackage{svg}
\usepackage{cite}

\makeatletter
\let\NAT@parse\undefined
\makeatother
\usepackage[colorlinks=true,linkcolor=red,citecolor=blue]{hyperref}

\title{\LARGE \bf
Skeleton-Based Human Action Recognition with Noisy Labels}

\author{Yi Xu$^{1}$, Kunyu Peng$^{1,*}$, Di Wen$^{1}$, Ruiping Liu$^{1}$, Junwei Zheng$^{1}$, Yufan Chen$^{1}$, Jiaming Zhang$^{1}$,\\Alina Roitberg$^{2}$, Kailun Yang$^{3,4}$, and Rainer Stiefelhagen$^{1}$%
\thanks{This work was supported in part by the SmartAge project sponsored by the Carl Zeiss Stiftung (P2019-01-003; 2021-2026), by the project served to prepare the SFB 1574 Circular Factory for the Perpetual Product (project ID: 471687386), approved by the Deutsche Forschungsgemeinschaft (DFG, German Research Foundation) with a start date of April 1, 2024, and in part by the BMBF through a fellowship within the IFI program of the German Academic Exchange Service (DAAD), in part by the HoreKA@KIT supercomputer partition, and in part by Hangzhou SurImage Technology Company Ltd. The authors also acknowledge support by the state of Baden-W\"urttemberg through bwHPC and the German Research Foundation (DFG) through grant INST 35/1597-1 FUGG.}
\thanks{*Corresponding author. (Email: kunyu.peng@kit.edu.)}%
\thanks{$^{1}$The authors are the Institute for Anthropomatics and Robotics, Karlsruhe Institute of Technology, Germany.}%
\thanks{$^{2}$The author is with the Institute for Artificial Intelligence, University of Stuttgart, Germany.}%
\thanks{$^{3}$The author is with the School of Robotics, Hunan University, China.}%
\thanks{$^{4}$The author is also with the National Engineering Research Center of Robot Visual Perception and Control Technology, Hunan University, China.}
}

\begin{document}

\maketitle
\thispagestyle{empty}
\pagestyle{empty}

\begin{abstract}
Understanding human actions from body poses is critical for assistive robots sharing space with humans in order to make informed and safe decisions about the next interaction. However, precise temporal localization and annotation of activity sequences is time-consuming and the resulting labels are often noisy. If not effectively addressed, label noise negatively affects the model's training, resulting in lower recognition quality. Despite its importance,  addressing label noise for skeleton-based action recognition has been overlooked so far. In this study, we bridge this gap by implementing a framework that augments well-established skeleton-based human action recognition methods with label-denoising strategies from various research areas to serve as the initial benchmark. Observations reveal that these baselines yield only marginal performance when dealing with sparse skeleton data. Consequently, we introduce a novel methodology, NoiseEraSAR, which integrates global sample selection, co-teaching, and Cross-Modal Mixture-of-Experts (CM-MOE) strategies, aimed at mitigating the adverse impacts of label noise. Our proposed approach demonstrates better performance on the established benchmark, setting new state-of-the-art standards. The source code for this study is accessible at \textit{\url{https://github.com/xuyizdby/NoiseEraSAR}}.
\end{abstract}

\section{Introduction}

Skeleton-based human action recognition is vital in robotics, particularly in human-robot interaction, enabling more natural and intuitive communication~\cite{dallel2020inhard,rodomagoulakis2016multimodal}. It plays a crucial role in surveillance~\cite{lee2020real,akkaladevi2015action}, allowing robots to identify and respond to emergencies or unsafe behaviors. Skeleton-based human action recognition also enables personalized services in areas like healthcare and fitness by analyzing and adapting to individual human movements~\cite{yu2021adaptive,cho2020self}. Additionally, it improves efficiency and safety in collaborative industrial environments by enabling robots to understand and predict human actions~\cite{pham2020spatio,koch2022methods}.
The success of deep learning for human action recognition heavily relies on data annotation for supervised training. However, the cost-effective and decentralized manual labeling work on crowdsourcing platforms introduces significant concerns related to the quality of the annotation~\cite{althnian2021impact}. 
Additionally, the development of large models necessitates diverse and extensive datasets with large amounts of labels. The quality of the labels will directly affect the model training~\cite{song2019selfie,wei2021smooth,chen2021beyond}. Due to the lack of visual appearance, skeleton data is hard to annotate compared with video data, which may result in more label noise.
Therefore, designing models that can alleviate the negative effect brought by label noise is a challenge for skeleton-based action recognition in real-world scenarios.
\begin{figure}[t!]
    \centering
    \includegraphics[width=0.5\textwidth]{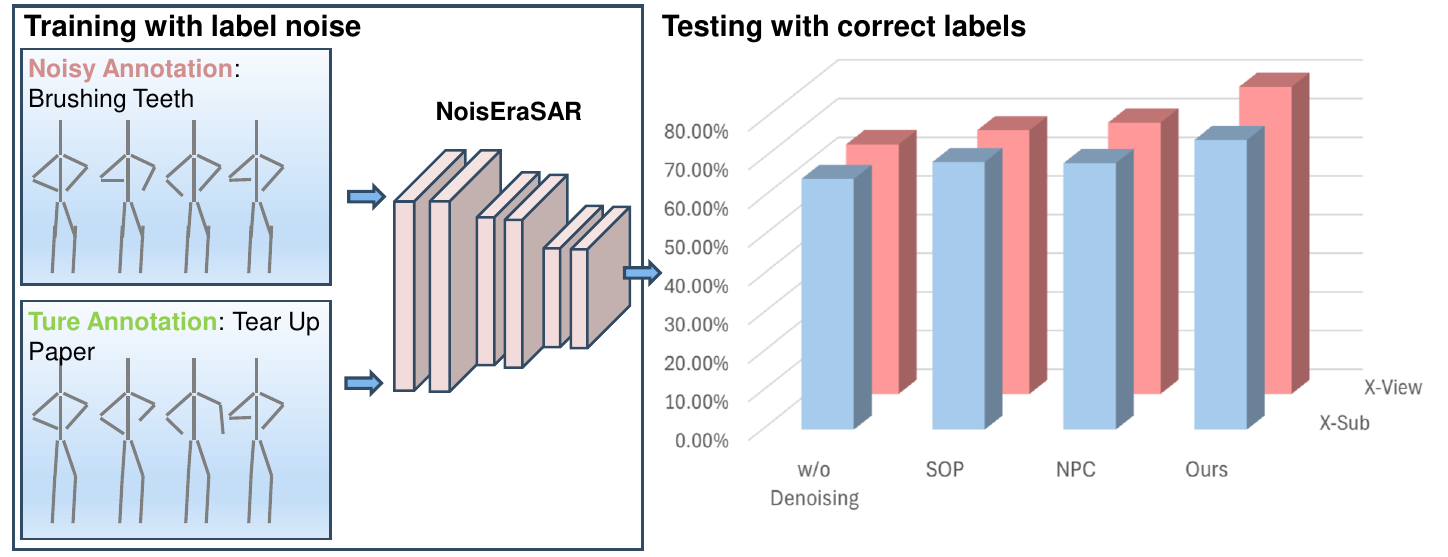}
    \vskip-2ex
    \caption{An overview of our task setting. We randomly inject asymmetric label noise into the training set according to a predefined noise ratio. On the right-hand side, we deliver the comparison of the performances on the test set with correct labels in terms of the Cross-Subject (X-Sub) and the Cross-View (X-View) settings, where our approach shows the best performance.}
    \label{teaser}
    \vskip-4ex
    \end{figure} 
In recent literature concerning the mitigation of label noise, various methodologies have been proposed, encompassing techniques such as sample selection and noise modeling~\cite{Han_Yao_Yu_Niu_Xu_Hu_Tsang_Sugiyama_2017,chen2019understanding,li2020dividemix}. These methodologies have been further extended by some researchers who integrate sample selection within a semi-supervised learning paradigm~\cite{Liu_Zhu_Qu_You}, categorizing clean annotated data as labeled and unclean annotated data as unlabeled, thereby leveraging the strengths of semi-supervised learning techniques. Moreover, a distinct subset of studies has focused on the explicit modeling of data labels, primarily through the construction of transition matrices, to systematically address the issue of label noise~\cite{Bae_Shin_Jang_Na_Song_Moon_2022}.
Nevertheless, the literature lacks investigations specifically targeting skeleton-based human action recognition in the context of noisy label scenarios, whereas the majority of existing studies are concentrated on image classification tasks.

The objective of this paper is to conduct the first work on skeleton-based action recognition under a noisy label setting and introduce a novel approach for mitigating label noise within this new task. 
To conduct the first testbed, two well-established label denoising approaches are implemented into skeleton-based action recognition pipeline, which are SOP~\cite{Liu_Zhu_Qu_You} and NPC~\cite{Bae_Shin_Jang_Na_Song_Moon_2022}, where we firstly choose CTR-GCN~\cite{chen2021channel} as backbone for skeleton feature learning and conduct experiments on different label noise ratios. We find these two label-denoising methods can not well address the label noise scenario for the skeleton-based action recognition task, which is mostly due to the sparsity of the skeleton data. We thereby propose a new approach in this field to handle the aforementioned challenges, which is named as NoiseEraSAR.
This approach synergizes methodologies pertaining to sample selection and multi-modality fusion. The sample selection mechanism reduces the propensity for model overfitting to noisy data by employing a co-selection strategy between two models. Concurrently, the multi-modality fusion aspect addresses the distinctive attributes of skeletal data, harnessing complementary insights from multi-modal sources to fortify training resilience against label noise. This is achieved through the new proposed Cross-Modal Mixture-of-Experts (CM-MoE) framework, leveraging a spatio-temporal graph gate network. Our proposed method achieves state-of-the-art performances on the constructed testbed, as shown in Figure~\ref{teaser}.

Our contributions are summarized as follows:
\begin{itemize}

\item We open the vistas for Skeleton-based Human Action Recognition (SHAR) under noisy labels and create a new benchmark by implementing the well-established methods from other domains, which focus on the label noise problem, into the skeleton-based action recognition field.

\item We introduce a new method, NoiseEraSAR, to address label noise in SHAR. It combines the principles of cross-training and the Cross-Modal Mixture-of-Experts (CM-MoE) to formulate an effective training framework for SHAR under noisy labels. We also conduct comprehensive ablation experiments to verify the effectiveness of different components of the method.

\item Experimental results demonstrate the effectiveness of the proposed method across different label noise ratios. NoiseEraSAR clearly surpasses state-of-the-art methods in various label noise settings. Our method achieves $74.9\%$ and $79.5\%$ of accuracy on cross-subject and cross-view evaluations on the NTU-60 dataset under $80\%$ label noise.

\end{itemize}

\section{Related Work}
\subsection{Skeleton-based Human Action Recognition}
Skeleton-based human action recognition is exceptionally suitable for various applications that demand high reliability and efficiency, 
due to its strong environmental robustness and extensive capability for subject generalization.

Early studies leaned towards CNN-based methods~\cite{wang2016action,soo2017interpretable,tu2018skeleton},
leveraging their superior 
hierarchical feature learning, alongside RNN-based methods~\cite{du2015hierarchical,zhang2019view, si2019attention}, 
recognized for 
modeling temporal dynamical behavior in sequences. Recent approaches have shifted to transformer-based methods~\cite{zhang2021stst,lee2023hierarchically,ahn2023star}, 
acclaimed in capturing long-range dependencies and facilitating parallel processing for efficiency. To adeptly utilize skeletal geometric information, GCN-based methods~\cite{yan2018spatial,ye2020dynamic,chen2021channel,peng2023navigating} 
focus on topology modeling as a fundamental design principle. 
ST-GCN~\cite{yan2018spatial}, which initially utilized fixed graph convolution, effectively models dynamic spatial-temporal relationships using a predefined skeleton topology.
Despite its efficiency, the reliance on a fixed graph structure introduces limitations in adaptability and precision for recognizing dynamic and diverse human actions.
Dynamic GCN~\cite{ye2020dynamic} introduces adaptive graph topologies that allow for more flexible and accurate modeling of complex human actions. CTR-GCN~\cite{chen2021channel} innovates by refining the graph topology at a channel-wise aspect, enabling more precise and adaptive capture of spatial-temporal relationships in human actions. 
We construct our main benchmark on CTR-GCN~\cite{chen2021channel} to achieve the backbone unification while conducting ablation for different skeleton-based action recognition backbones on HD-GCN~\cite{lee2023hierarchically} and ST-GCN~\cite{yan2018spatial}.

\subsection{Noisy Labels Learning}
Noisy labels learning focuses on efficiently training
models on datasets with inaccurate labels, aiming to enhance model robustness and accuracy despite the presence of label noise.

Strategies to enhance the robustness of models 
can be roughly
classified into five categories~\cite{song2022learning}, \textit{i.e.}, (i) Modifying loss for noisy labels, 
which can be further achieved through four aspects: (a) Estimating the noise label transition matrix to adjust the loss function~\cite{xia2019anchor,tanno2019learning,zhu2021clusterability,zhu2022beyond,li2022estimating};
(b) Re-weighting individual sample losses by reducing the weight of samples likely mislabeled~\cite{liu2015classification};
(c) Refurbishing the labels that are presumed noisy~\cite{song2019selfie,wei2021smooth,chen2021beyond};
(d) Adjusting with the optimal rule deriving from meta-learning~\cite{shu2019meta,wang2020training,zheng2021meta}.
(ii) Developing loss functions robustness to noise~\cite{liu2020peer,ma2020normalized,zhu2021second}:
These are aiming to inherently tolerate label inaccuracies without requiring a noise transition matrix. 
(iii) Applying regularization methods~\cite{wei2021open,cheng2021mitigating,liu2022robust}: 
These methods aim to leverage regularization techniques to prevent over-fitting to noisy labels, thereby enhancing the overall robustness and generalization of the models.
(iv) Enhancing robustness with architecture~\cite{cheng2020weakly}:
Integrating a specialized noise adaptation structure atop the base Deep Neural Network (DNN) facilitates understanding of the label transition mechanism, thereby enhancing training robustness against the label noise.
(v) Implementing dynamic sample selection~\cite{Han_Yao_Yu_Niu_Xu_Hu_Tsang_Sugiyama_2017,chen2019understanding,li2020dividemix}: 
This approach starts with selecting a subset of clean samples and progressively incorporates incorrectly labeled samples in a semi-supervised learning framework.
Notably, our model uses a cross-training scheme inspired by~\cite{Han_Yao_Yu_Niu_Xu_Hu_Tsang_Sugiyama_2017} to identify small-loss (and thus likely ``clean'') samples for further training.

\section{Benchmark}
\subsection{Label Noise}

In our benchmark, we establish an initial testing framework specifically designed for skeleton-based human action recognition, incorporating the challenge of noisy labels. Our goal is to assess how well a model can perform under various levels of label inaccuracies. This involves deliberately mislabeling a certain percentage of the samples while making sure these incorrect labels are still among those available in the dataset utilized.
We distinguish between two main types of noise: \textbf{Symmetric noise} and \textbf{Asymmetric noise}. \textbf{Symmetric noise} is characterized by the random alteration of each label to any other class with a uniform probability, as described in \cite{patel2023adaptive}. In contrast, \textbf{Asymmetric noise} involves changing labels to different classes based on distinct probabilities. For example, if categories A and B are typically more confusable, we might specify that a mislabeled sample from class A has a $0.6$ probability of being wrongly assigned to class B and a $0.4$ chance of being attributed to any other category.

Implementing \textbf{Asymmetric noise} necessitates a pre-established understanding of the likelihood of confusion between specific categories, informed by real-world data labeling experiences. 
In this study, our focus is restricted to scenarios with \textbf{Symmetric noise}. Studying symmetric label noise is crucial as it provides a foundational understanding of how random errors in labeling affect model training, setting the stage for tackling the more complex and realistic scenarios of asymmetric label noise. 
We have introduced symmetric noise into the training set at various levels: $20\%$, $40\%$, $50\%$, and $80\%$. To ensure the integrity of our evaluation, we do not introduce label noise into the testing set.

\subsection{Baseline}
A crucial aspect of our benchmark is the selection of the baseline methods. We opt for two established techniques as our baselines due to their proven effectiveness in handling noisy labels across different domains. These methods serve as a standard for comparing the performance of our model, and we conduct experiments based on them. We choose CTR-GCN~\cite{chen2021channel} as the feature extraction backbone to construct our main test bed for skeleton-based human action recognition under different label noise ratios while leveraging the other two backbones, \textit{i.e.}, HD-GCN~\cite{lee2023hierarchically} and ST-GCN~\cite{yan2018spatial}, for the evaluation of the cross-backbone generalizability of different denoising methods.

\noindent\textbf{Sparse Over-Parameterization (SOP)}~\cite{Liu_Zhu_Qu_You}\textbf{.}
SOP is an approach for robust training against label noise, which models the label noise via another sparse over-parameterization term and exploits implicit algorithmic regularizations to separate the noise. 
\noindent\textbf{Noisy Prediction Calibration (NPC)}~\cite{Bae_Shin_Jang_Na_Song_Moon_2022}\textbf{.}
NPC estimates the explicit transition from a noisy prediction to a true latent class via utilizing a deep generative model. 

\subsection{Dataset}
\textbf{NTU RGB+D.} NTU RGB+D (NTU-60)~\cite{Shahroudy_Li_Ng_Wang_2016} is a large 3D human action recognition dataset, which includes $56,880$ samples covering a total of $60$ action classes. Each sample is captured from three different viewpoints, including one action type and up to two subjects. The dataset comprises various modalities, including skeleton, IR, and RGB. For our experiments, we select the skeleton temporal modality, which involves a total of $25$ joint points. The 3D coordinates of each joint point are recorded for each frame as the original features. The NTU-60 dataset is a popular choice for skeleton-based human action recognition problems and is therefore selected as the primary dataset for our experiments.

\section{Methodology}

\subsection{Overview}
\begin{figure*}[htp]
    \centering
    \includegraphics[width=1.0\textwidth]{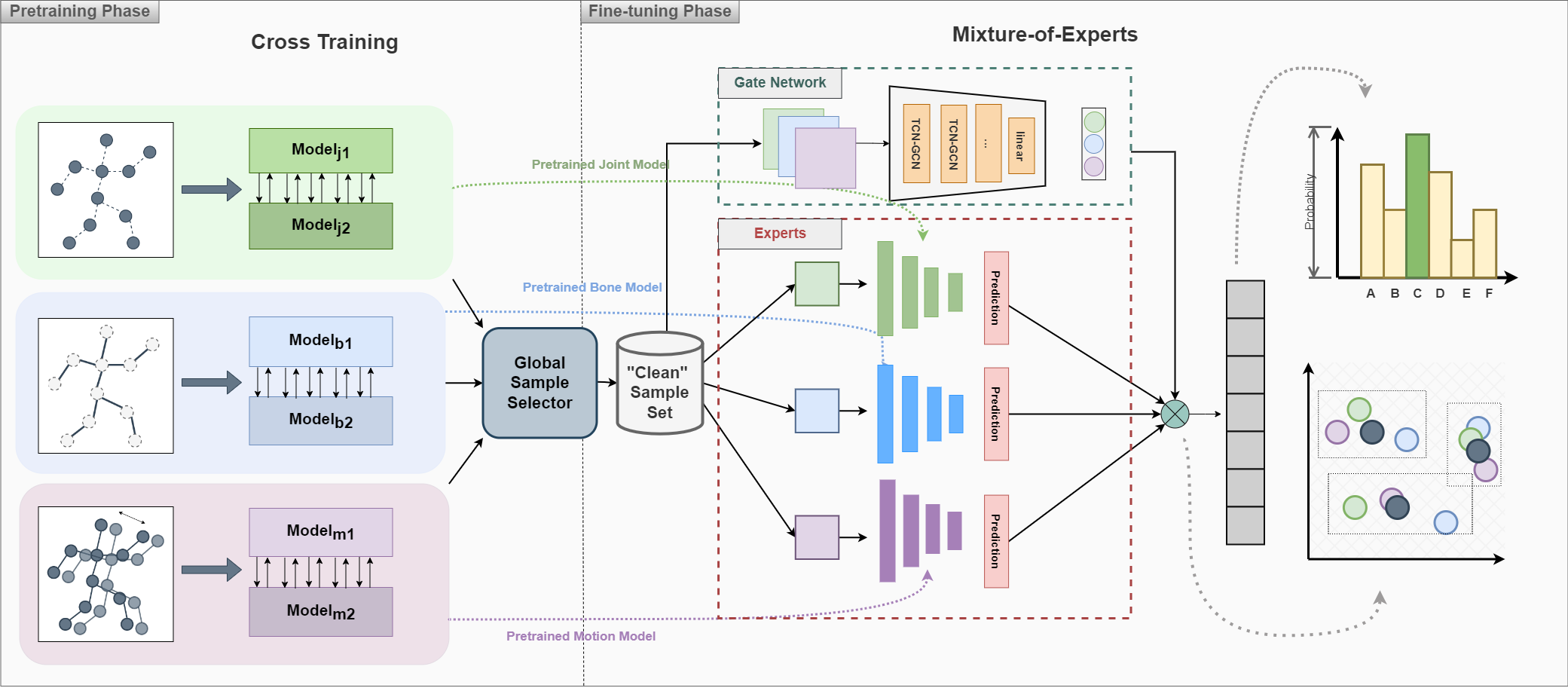}
    \vskip-2ex
    \caption{Overview of the method, NoiseEraSAR: In the pre-training phase, the proposed method first trains special models for joint, bone, and motion modalities by using a cross training method. The small clean dataset is generated from the pre-trained models by evaluating the loss value, and it is fed into the Cross-Modal Mixture-of-Experts (CM-MoE). In the fine-tuning phase, the gate network is added to control the weights of each expert and assists the CM-MoE.}
    \vskip-4ex
    \label{structure}
    \end{figure*}

We first deliver an overview of the contributed new method in this subsection.
As illustrated in Figure \ref{structure}, our method, NoiseEraSAR, contains two main phases: The pretraining phase and the fine-tuning phase. 
Especially, three components are involved: cross-training, global sample selection, and a Cross-Modal Mixture of Expert (CM-MoE) technique. Firstly, data from three modalities (joint, bone, and motion represented in green, blue, and purple) are extracted for each sample. 
For each modality, two models with the same structure are simultaneously trained using the co-teaching method, where these models select samples whose labels are worthy of belief according to the losses for each other in every epoch (\textit{e.g.}, ${Model}_{j1}$ and ${Model}_{j2}$ in Figure~\ref{structure} are the peer networks of the joint modality).
Next, in the global sample selection process, models from the three modalities individually choose a defined percentage of clean labeled samples from the whole training dataset and the final sample set is constructed by taking the union of these three selected sets considering the samples.
The final step involves employing the CM-MoE system while relying on the union of the samples from the selected three clean labeled sets. 
The triple networks pre-trained in the first step will be combined with a gate network constructed by spatiotemporal graph convolution that connects the starting input layer and the SoftMax layer (as shown on the right side of Figure~\ref{structure}). The gate network dynamically adjusts the weights of the outputs from different models across modalities. The weighted average of the three scores will be used as the prediction of our final method. Below, we present a detailed overview of the different components comprising our method.

\subsection{Modalities}
We leverage three skeletal modalities: joint, bone, and motion. Both bone and motion modalities are derived from the original joint coordinate data.

\noindent\textbf{Joint modality.}
Joint modality is the raw form of skeletal data, which mainly records the 3D coordinates ($\mathbf{x}$, $\mathbf{y}$, and $\mathbf{z}$) of each joint point at different times (frames).  In particular, the NTU-60 dataset contains a total of $25$ joints with their coordinates at each frame. Then we could describe the joint data of joint index $i$ at time $t$ as:
\begin{align}
j_{i, t}=\left(x_{i, t}, y_{i, t}, z_{i, t}\right)
\end{align}

\noindent\textbf{Bone modality.}
A bone vector can be described by the head and tail joint points. For example, while processing with the NTU-60 dataset, we calculate a total of $25$ bones (\textit{i.e.} $25$ joint pairs). Therefore, we can simply calculate the bone data ($b_{i, j}$) based on the joint pairs($x_{j, t}$ and $x_{i, t}$):
\begin{align}
b_{i, j}=\left(x_{j, t}-x_{i, t}, y_{j, t}-y_{i, t}, z_{j, t}-z_{i, t}\right)
\end{align}

\noindent\textbf{Motion modality.}
The motion modality is described by the difference between the same joints in two consecutive time points (frames). The motion data of joint $i$ between the time $t$ and $t+1$ can be represented as follows: 
\begin{align}
m_{i, t, t+1}=\left(x_{i, t+1}-x_{i, t}, y_{i, t+1}-y_{i, t}, z_{i, t+1}-z_{i, t}\right)
\end{align}

\subsection{Cross Training} 
The first part of the method is the cross-training process under three different modalities (joint, bone, and motion). Inspired by the method used in co-training~\cite{Han_Yao_Yu_Niu_Xu_Hu_Tsang_Sugiyama_2017}, we simultaneously trains two networks ($\mathbf{N}_{1}$ and $\mathbf{N}_{2}$) with identical structure. The small-loss method is applied here, where a certain proportion of samples with smaller losses within each batch is identified as clean samples at each epoch, and only these samples participate in the back-propagation process. Consequently, at each epoch, both networks select samples for their counterpart based on the small-loss criterion.

If in each batch of the epoch ($T$), we have the mini-batch dataset $\overline{D}$, The process of sample selection in each epoch can be described as follows:
\begin{align}
R(T)=1-\min \left\{\frac{T}{T_{in}} r, r\right\}
\end{align}
\begin{align}
\overline{{D}}_{1}={\arg \min}_{{D}^{\prime}:\left|{D}^{\prime}\right| \geq R(T)|\overline{{D}}|} \ell\left({\mathbf{N}}_{1}, {D}^{\prime}\right)
\end{align}
\begin{align}
\overline{{D}}_{2}={\arg \min} _{{D}^{\prime}:\left|{D}^{\prime}\right| \geq R(T)|\overline{{D}}|} \ell\left({\mathbf{N}}_{2}, {D}^{\prime}\right)
\end{align}
where $R(T)$ is the the ratio of selected samples and $r$ is the noise ratio. And the $T_{in}$ represents the fixed number of epochs and $\ell$ is the cross-entropy loss function. Finally, we obtain two set of samples ($\overline{{D}}_{1}$ and $\overline{{D}}_{2}$)

Then, each network updates its parameters by the selected samples from the peer as follows:
\begin{align}
w_{1}=w_{1}-\eta \nabla \ell\left({\mathbf{N}}_{1}, \overline{{D}}_{2}\right)
\end{align}
\begin{align}
w_{2}=w_{2}-\eta \nabla \ell\left({\mathbf{N}}_{2}, \overline{{D}}_{1}\right)
\end{align}
where $w_{1}$ and $w_{2}$ represent the trainable parameters of the $\mathbf{N}_{1}$ and $\mathbf{N}_{2}$. $\mathbf{N}_{1}$ represents $Model_{\phi1}$ and $\mathbf{N}_{2}$ represents $Model_{\phi2}$, while $\phi \in \left[j,b,m\right]$, as shown in Figure~\ref{structure}.

Throughout the cross-training process, the tendency to memorize noisy labels is mitigated, since the peer networks have different learning abilities and can filter out different types of introduced errors by noisy labels.

After the last epoch, the network with superior accuracy is selected as the final model for this modality. 
Therefore, we obtain three pre-trained networks  ($\mathbf{N}_{joint}$, $\mathbf{N}_{bone}$, and $\mathbf{N}_{motion}$).

\subsection{Global Sample Selection}
After the pre-training process with the cross-training method for each modality, a global sample selector is proposed to extract more representative samples for the next modality fusion process. 
Given that the entire dataset includes a significant proportion of noisy labels, sample selection proves beneficial for further training, both in terms of denoising and improving calculation speed. 
If the ratio of instances to be selected is set to ${p}$ and training set ${D}$ is given, then we have the selected samples from three networks as follows:
\begin{align}
{D}_{joint}={\arg \min} _{{D}^{\prime}:\left|{D}^{\prime}\right| \geq p|{D}|} \ell\left(\mathbf{N}_{joint}, {D}^{\prime}\right)
\end{align}
\begin{align}
{D}_{bone}={\arg \min} _{{D}^{\prime}:\left|{D}^{\prime}\right| \geq p|{D}|} \ell\left(\mathbf{N}_{bone}, {D}^{\prime}\right)
\end{align}
\begin{align}
{D}_{motion}={\arg \min} _{{D}^{\prime}:\left|{D}^{\prime}\right| \geq p|{D}|} \ell\left(\mathbf{N}_{motion}, {D}^{\prime}\right)
\end{align}
To obtain the benefits of triple modalities and to avoid the oversampling of the examples with lower learning difficulty from single networks, the final clean set $D_c$ is the global union set of samples from three networks:
\begin{align}
{D}_{c}={D}_{joint}\cup{D}_{bone}\cup{D}_{motion}
\end{align}

\subsection{Cross-Modal Mixture of Expert (CM-MoE)}
\begin{table*}[t!]
\caption{The comparison results of skeleton-based human action recognition on the NTU-60~\cite{Shahroudy_Li_Ng_Wang_2016} dataset.}
\vskip-3ex
\label{tab:result}
\renewcommand\arraystretch{1.2}
\begin{center}
\resizebox{0.9\linewidth}{!}{%
\begin{tabular}{p{3.1cm}|p{1.3cm}|p{1.3cm}|p{1.3cm}|p{1.3cm}|p{1.3cm}|p{1.3cm}|p{1.3cm}|p{1.3cm}}
\hline
\multicolumn{1}{l|}{\textbf{}} &
  \multicolumn{8}{c}{\textbf{Symmetric Noise}} \\ \cline{2-9}
\multirow{2}{*}{\textbf{Method/Noise ratio}} &
  \multicolumn{2}{c|}{\textbf{20\%}} &
  \multicolumn{2}{c|}{\textbf{40\%}} &
  \multicolumn{2}{c|}{\textbf{50\%}} &
  \multicolumn{2}{c}{\textbf{80\%}} \\ \cline{2-9} 
 &
  \multicolumn{1}{l|}{X-Sub} &
  \multicolumn{1}{l|}{X-View} &
  \multicolumn{1}{l|}{X-Sub} &
  \multicolumn{1}{l|}{X-View} &
  \multicolumn{1}{l|}{X-Sub} &
  \multicolumn{1}{l|}{X-View} &
  \multicolumn{1}{l|}{X-Sub} &
  X-View \\ \hline
CTR-GCN~\cite{2021Channel} &
  \multicolumn{1}{l|}{\underline{86.8\%}} &
  \multicolumn{1}{l|}{\underline{90.9\%}} &
  \multicolumn{1}{l|}{\underline{83.7\%}} &
  \multicolumn{1}{l|}{\underline{88.2\%}} &
  \multicolumn{1}{l|}{81.7\%} &
  \multicolumn{1}{l|}{\underline{85.7\%}} &
  \multicolumn{1}{l|}{64.8\%} &
  64.6\% \\
CTR-GCN~\cite{2021Channel} + SOP~\cite{Liu_Zhu_Qu_You} &
  \multicolumn{1}{l|}{83.7\%} &
  \multicolumn{1}{l|}{88.3\%} &
  \multicolumn{1}{l|}{82.0\%} &
  \multicolumn{1}{l|}{87.3\%} &
  \multicolumn{1}{l|}{81.0\%} &
  \multicolumn{1}{l|}{85.6\%} &
  \multicolumn{1}{l|}{\underline{69.2\%}} &
  68.2\% \\
CTR-GCN~\cite{2021Channel} + NPC~\cite{Bae_Shin_Jang_Na_Song_Moon_2022} &
  \multicolumn{1}{l|}{86.6\%} &
  \multicolumn{1}{l|}{90.8\%} &
  \multicolumn{1}{l|}{83.2\%} &
  \multicolumn{1}{l|}{87.4\%} &
  \multicolumn{1}{l|}{\underline{82.4\%}} &
  \multicolumn{1}{l|}{85.5\%} &
  \multicolumn{1}{l|}{68.9\%} &
  \underline{70.2\%} \\ \hline
\multicolumn{1}{l|}{NoiseEraSAR (ours)} &
  \multicolumn{1}{l|}{\textbf{90.6\%}} &
  \multicolumn{1}{l|}{\textbf{95.3\%}} &
  \multicolumn{1}{l|}{\textbf{89.0\%}} &
  \multicolumn{1}{l|}{\textbf{93.1\%}} &
  \multicolumn{1}{l|}{\textbf{88.5\%}} &
  \multicolumn{1}{l|}{\textbf{90.5\%}} &
  \multicolumn{1}{l|}{\textbf{74.9\%}} &
  \textbf{79.5\%} \\ \hline
\end{tabular}%
}
\end{center}
\vskip-4ex
\end{table*}
\begin{table*}[t!]\tiny%
\caption{Ablation study of the proposed components of our method, including cross-training, global sample selector, and Cross-Modal Mixture-of-Experts technique (CM-MoE). }
\vskip-4ex
\label{tab:ablation}
\renewcommand\arraystretch{1.2}
\begin{center}
\resizebox{0.8\linewidth}{!}{%
\begin{tabular}{l|l|c|c}
\hline
\multirow{2}{*}{\textbf{Methods}}         & \multirow{2}{*}{\textbf{Modality}} & \multicolumn{2}{c}{\textbf{Symmetric Noise 80\%}} \\ \cline{3-4}
                                &                   & {\textbf{X-Sub}} & \textbf{X-View} \\ \hline
CTR-GCN\cite{2021Channel}+Cross-training          & joint             & 70.8\%                              & 74.9\%          \\
CTR-GCN\cite{2021Channel}+Cross-training                                & bone              & 65.2\%                              & 69.0\%          \\
CTR-GCN\cite{2021Channel}+Cross-training                                & motion            & 67.5\%                              & 69.6\%          \\
\hline
CTR-GCN\cite{2021Channel}+Cross-training+ensemble & joint+bone+motion & 72.5\%                              & 78.9\%          \\ \hline
\textbf{CTR-GCN}\cite{2021Channel}\textbf{+Cross-training+CM-MoE (ours)} & joint+bone+motion                  & \textbf{74.9\%}         & \textbf{79.5\%}         \\ \hline
\end{tabular}%
}
\end{center}
\vskip-5ex
\end{table*}

In the domain of skeleton-based human action recognition, models trained on joint, bone, and motion modalities learn complementary patterns from the data, even when using identical the same network structures. Therefore, they are expected to provide complementary information to achieve label denoising. 
Consequently, we introduce the concept of the CM-MoE technique for fusing models of different modalities to acquire robust embeddings against label noise. The gate structure in such a system can dynamically adjust the importance of each modality for a specific action, thereby enhancing the robustness of the model in noisy environments.

After obtaining three pre-trained models and a dataset with labels worthy of belief, we construct a spatiotemporal graph-based gate network structure to control the varying weights of the SoftMax layer of the three models. 
The gate network receives raw data compiled from the three modalities, with the three pre-trained models serving as experts to process their respective modalities.

Let ${S}^{joint}_{c}= \mathbf{N}_{joint}\left(\mathbf{j}_{c}\right)$, ${S}^{bone}_{c}= \mathbf{N}_{bone}\left(\mathbf{b}_{c}\right)$ and ${S}^{motion}_{c}= \mathbf{N}_{motion}\left(\mathbf{m}_{c}\right)$ as the SoftMax score of each expert network, where $\mathbf{j}_{c}$, $\mathbf{b}_{c}$, and $\mathbf{m}_{c}$ are derived from one sample in $D_c$.
The final prediction result is calculated as follows:
\begin{align}
\mathbf{W}= \mathbf{G}_{\theta}\left(\mathbf{j}_{c}||\mathbf{b}_{c}||\mathbf{m}_{c}\right)
\end{align}
\begin{align}
{S} = \mathbf{W}\left({S}^{joint}_{c}||{S}^{bone}_{c}||{S}^{motion}_{c}\right)
\end{align}
where $\mathbf{G}_{\theta}$, $W$, and $S$ represent the gate network, the output (weights for three expert networks) of the gate network, and the final prediction score. $||$ represents the concatenation operation.
Specifically, The gate network $\mathbf{G}_{\theta}$
comprises two basic blocks in CTR-GCN~\cite{chen2021channel}, followed by global average pooling and a SoftMax classifier to generate the predicted weights for each expert.

\section{Experiments}

\subsection{Experimental Settings}
All experiments in this section are implemented on RTX 6000 GPU with the PyTorch deep learning framework. And all data pre-processing methods follow CTR-GCN~\cite{2021Channel}. 

\noindent\textbf{Backbone.} For a fairer performance comparison, we implement the CTR-GCN~\cite{chen2021channel} model as the backbone of the NPC and SOP methods in our testbed. Additionally, we select HD-GCN~\cite{lee2023hierarchically} and ST-GCN~\cite{yan2018spatial} for ablation of the backbones.

\noindent\textbf{Hyperparameters.} 
In the cross-training part, the hyperparameters setting follows CTR-GCN. each model is trained using SGD with a momentum of $0.9$ and weight decay of $0.0004$. The basic learning rate is set to $0.1$ and the training epoch is set to $65$. In the part of fine-tuning with the CM-MoE network, and weight decay is set to $0.0005$, the basic learning rate and the training epoch are set to $0.1$ and $10$.

\begin{figure*}[htp]
    \centering
    \includegraphics[width=1.0\textwidth]{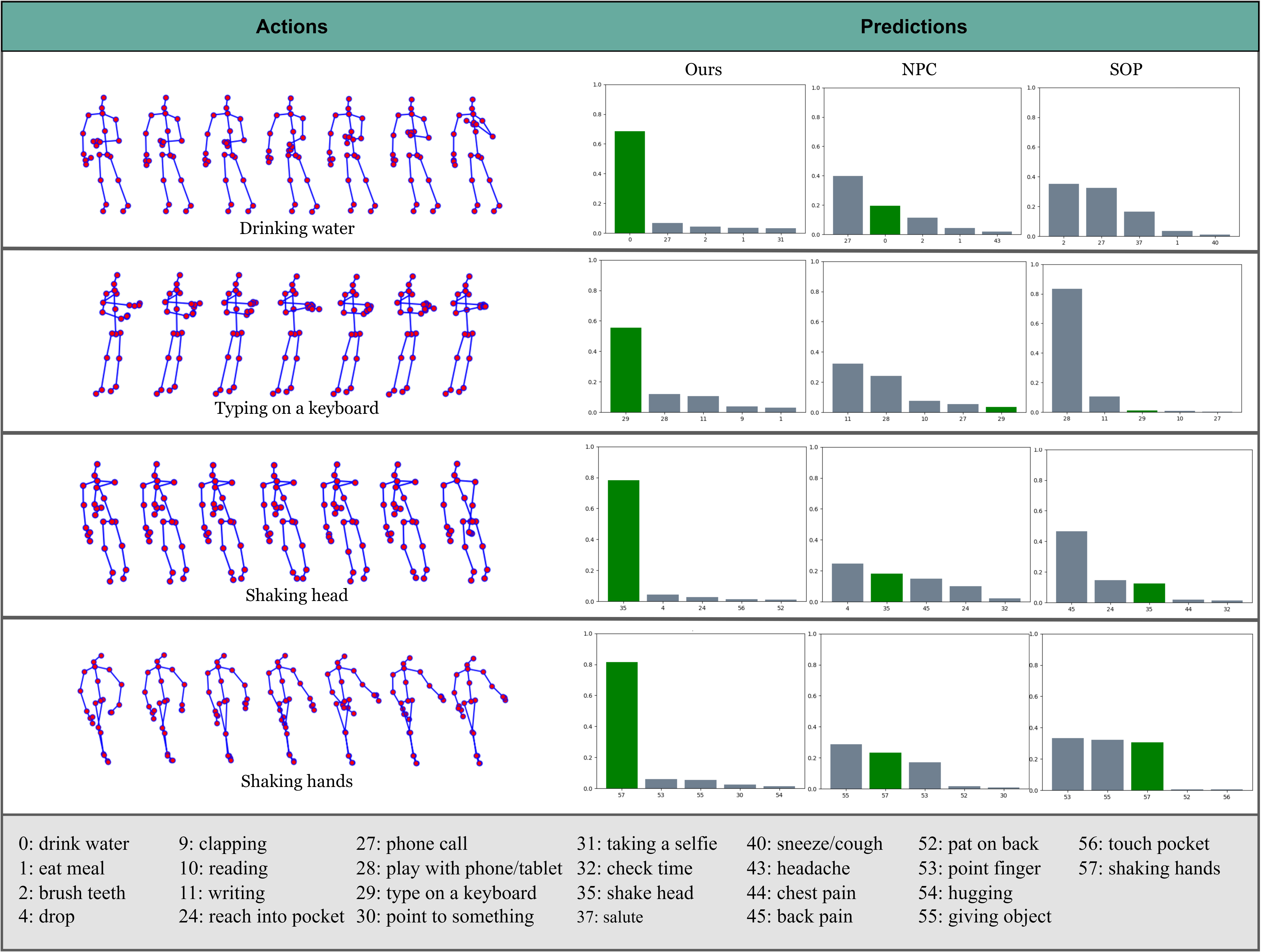}
    \vskip-2ex
    \caption{Prediction results of the three methods. Four samples from different action classes are visualized on the left. The Top-5 SoftMax scores are drawn on the right with the ground truth (green color). All these predictions are generated from the models under the $80\%$ Cross-View setting.}
    \label{qualitative}
    \end{figure*} 

\subsection{Results}
\label{sec:results}
In this section, we compare the performance of our method with two typical existing methods (NPC~\cite{Bae_Shin_Jang_Na_Song_Moon_2022} and SOP~\cite{Liu_Zhu_Qu_You}), which focus on the noisy label problem in the field of image classification. Table~\ref{tab:result} showcases our method's performance, subjected to symmetric noise levels ranging from $20\%$ to $80\%$ on the NTU-60~\cite{Shahroudy_Li_Ng_Wang_2016} dataset while using CTR-GCN~\cite{chen2021channel} as the backbone. We randomly choose $4$ different noise ratios among $0.0\%$ to $100.0\%$ to set up this benchmark, encompassing easy, middle, and hard levels.

\begin{table*}[htp]\tiny%
\caption{The Experiment with different backbones under $80\%$ symmetric noise, comparing with SOP~\cite{Liu_Zhu_Qu_You} method. }
\vskip-4ex
\label{tab:backbone}
\renewcommand\arraystretch{1.2}
\begin{center}
\resizebox{0.8\linewidth}{!}{%
\begin{tabular}{l|llllll}
\hline
\textbf{}                        & \multicolumn{6}{c}{Symmetric Noise 80\%}                                                                          \\ \cline{2-7} 
\multirow{2}{*}{Method/Backbone} & \multicolumn{2}{c|}{CTR-GCN \cite{{2021Channel}}}         & \multicolumn{2}{c|}{HD-GCN\cite{lee2023hierarchically}}                   & \multicolumn{2}{c}{ST-GCN\cite{yan2018spatial}} \\ \cline{2-7} 
                  & X-Sub  & \multicolumn{1}{l|}{X-View} & \multicolumn{1}{l|}{X-Sub} & \multicolumn{1}{l|}{X-View} & \multicolumn{1}{l|}{X-Sub} & X-View \\ \hline
SOP                              & \multicolumn{1}{l|}{69.2\%} & \multicolumn{1}{l|}{68.2\%} & \multicolumn{1}{l|}{71.1\%} & \multicolumn{1}{l|}{70.5\%} & \multicolumn{1}{l|}{47.5\%}  &   46.2\%\\ \hline
NoiseEraSAR (ours) & \multicolumn{1}{l|}{\textbf{74.9\%}} & \multicolumn{1}{l|}{\textbf{79.5\%}} & \multicolumn{1}{l|}{\textbf{76.2\%}}      & \multicolumn{1}{l|}{\textbf{76.5\%}}       & \multicolumn{1}{l|}{\textbf{54.6\%}}      &        \textbf{47.2\%}\\ \hline
\end{tabular}%
}
\end{center}
\vskip-7ex
\end{table*}

\noindent\textbf{Comparison with the State-of-the-Art.}
SOP and NPC demonstrate varying performance improvements over CTR-GCN without label denoising across different noise levels. In high-noise situations, both SOP and NPC achieve similar accuracies around $70\%$. At noise ratio from $20\%$ to $50\%$, NPC shows better performance than SOP, especially at $20\%$ where it can reach $92\%$ accuracy.

Our method consistently outperforms SOP and NPC across various noise levels. At the noise level of $80\%$, our model shows an improvement from $5.7\%$ to $10.3\%$ over the baselines. It indicates that, while SOP and NPC play a limited role in enhancing robustness at high noise levels, our method obtains further significant improvement. Even in scenarios with lower noise levels, our method surpasses SOP and NPC. For example, at $20\%$ noise, our method achieves an enhancement of approximately $5\%$. 

The main idea behind SOP and NPC is to model noisy labels, establishing a relationship between noisy labels and true labels. However, this approach is challenging to accurately estimate, particularly when dealing with a large number of classes~\cite{Han_Yao_Yu_Niu_Xu_Hu_Tsang_Sugiyama_2017}, and it is highly influenced by the modeling method. Regarding SHAR, datasets often have many categories, with high similarity between many of them, making the applicability of such methods relatively limited.

\noindent\textbf{Robustness of backbone.} 
Then, we analyze the CTR-GCN model's robustness without any additional anti-noise methods. Our experiments indicate that the CTR-GCN model remains sufficient robustness under $20\%$ to $50\%$ noise ratio, with test accuracy dropping from $86.8\%$ to $81.7\%$ (cross-subject) and from $90.9\%$ to $85.7\%$ (cross-view). However, at $80\%$ noise, a significant accuracy decline is evident, with $64.8\%$ (cross-subject) and $64.6\%$ (cross-view).
This highlights that the SHAR algorithm, exemplified by CTR-GCN, is sensitive to noisy labels since the powerful learning ability of these models comes with a great tendency to memorize random noise from the labels.

\noindent\textbf{Performance under different noise ratios.} 
Our method demonstrates superior performance in all noise environments. Notably, our model attains over $90\%$ top-1 accuracy in low-noise scenarios. For instance, at $20\%$, the method achieves $95.3\%$ accuracy at cross-view and $90.6\%$ accuracy at cross-subject. Furthermore, in high-noise environments (at $80\%$ noise ratio), the model performs well, reaching $74.9\%$ accuracy at cross-subject and $79.5\%$ at cross-view. Additionally, at $40\%$ and $50\%$ noise, the model maintains an accuracy of approximately $90\%$.
With the model's improved expressive capability during training, its denoising ability consistently strengthens through cross-training. Consequently, the model can adaptively obtain suitable denoising capabilities. Furthermore, our multi-modality fusion approach enhances the model's robustness by leveraging the complementarity among different modalities, thus mitigating excessive memorization of noisy labels within a single modality.
These findings underscore our method's adaptability to diverse noise environments, highlighting its potential for addressing real-world noise challenges.

\subsection{Qualitative Results}
\label{sec:qualitative}

To further assess our method against two other approaches, we select samples from four action classes that are easily confused by other categories when recognized by the human eye: \textit{Drinking water}, \textit{Typing on a keyboard}, \textit{Shaking hands}, and \textit{Shaking head}, as shown in Figure~\ref{qualitative}. 
Both SOP and NPC methods display less discriminative power, often ranking the correct class within the top $5$ but confusing it with similar classes—for example, NPC mixes up \textit{phone call} with \textit{drinking water}.

In contrast, our model excels in distinguishing different actions, consistently ranking the correct class higher than others, and showcasing improved confidence and accuracy. This enhancement is credited to our model's ability to effectively reduce noise in training and to utilize the complementary strengths of various modalities, such as joint and bone movements, to better understand and differentiate actions.
\subsection{Ablation Study}
In this section, we first conduct an ablation study to demonstrate the effectiveness of the key components of our NoiseEraSAR method. We ablate each component to quantify its contribution to the overall performance on NTU-60 with $80\%$ noisy labels. The result of the ablation study is shown in Table~\ref{tab:ablation}. Finally, we conduct the ablation to evaluate backbone generalizability.

\noindent\textbf{Cross Training.} 
First, comparing the CTR-GCN with other state-of-the-art methods in Sec.~\ref{sec:results}, it is observed that the cross-training method can effectively alleviate the issue of model performance degradation under a high proportion of label noise. With up to $80\%$ symmetric noise, the accuracy of model recognition under X-Sub improves from $64.8\%$ to $70.8\%$ (X-View improves from $64.6\%$ to $74.9\%$). The effectiveness of the cross-training approach for the SHAR under the noisy label problem is evident. The models cross-select samples with each other, which weakens the effect of over-memorization of noise.

\noindent\textbf{Multi-Modality Fusion.}
In our ablation experiments, we verify that the multi-modality fusion approach enhances model performance under label noise. We fuse the models after cross-training in three modalities, using a simple ensemble approach similar to CTR-GCN. Specifically, we calculate the weighted sum of the SoftMax layer outputs from the three model outputs. The weights selected for joint, bone, and motion are $0.6$, $0.6$, and $0.4$. 
The results show that although the recognition accuracy of the bone and motion modalities after cross-training is lower than that of joint modality ($70.8\%$ and $74.9\%$), the model's performance can be improved by $2\%$ to $4\%$ ($72.5\%$ and $78.9\%$) from the single modality. Therefore, we believe that models with different modalities have complementarity in human action recognition tasks after cross-training, and that modal fusion is significant for performance improvement.

\noindent\textbf{CM-MoE.}
We then conduct experiments to evaluate the effectiveness of the CM-MoE system. If the global sample selector and the CM-MoE system are not beneficial to address the issue of noisy labels, we would expect the utilization of these components can not improve the accuracy, compared with the simple ensemble method. Specifically, the model achieved a performance of $74.9\%$ under X-Sub and $79.5\%$ under X-View, surpassing the performance of the simple ensemble method (summation with fixed weights as presented in the previous section). In cases of a high ratio of noise, the MoE system can optimize the fusion of different modalities by assigning different weights to different features of the sample through the gate network. 

\noindent\textbf{Backbone Generalizability.}
We also conduct experiments with various backbones to assess the cross-backbone generalizability of our method. As shown in Table~\ref{tab:backbone}, our model consistently outperforms the SOP method on CTR-GCN~\cite{2021Channel}, HD-GCN~\cite{lee2023hierarchically}, and ST-GCN~\cite{yan2018spatial}. For instance, when employing HD-GCN as the backbone, the model achieves a recognition accuracy exceeding $76.0\%$, surpassing SOP by more than $5\%$. Even with ST-GCN, where both methods exhibit lower accuracy due to ST-GCN's weaker robustness in noisy environments, our approach still maintains an advantage over SOP. This result demonstrates the promising adaptability of our method to current state-of-the-art human action recognition algorithms when using them as backbones and validates the effectiveness of our proposed NoiseEraSAR independently of the backbone.

\section{Conclusion}
In this paper, we explore skeleton-based human action recognition with noisy labels. We created a benchmark using three label-denoising methods applied to GCN backbones. Our new method, NoiseEraSAR, employs co-teaching for multi-modal streams, global sample selection, and CM-MOE. NoiseEraSAR shows promising results on the NTU-60 dataset across various noisy label ratios and evaluation methods. This work aims to enhance robust skeleton-based action recognition models for robot-assisted human activities, even with low-quality training data, benefiting the community.

\bibliographystyle{IEEEtran}
\bibliography{bib}

\end{document}